\documentclass[pmlr,twocolumn,10pt]{jmlr}
\usepackage{booktabs}
\usepackage{longtable}
\usepackage{multirow}
\jmlrpages{}
\title[Flexible EHR Transformer]{Multi-Task Prediction of Clinical Outcomes in the Intensive Care Unit using Flexible Multimodal Transformers}
\author{
\Name{Benjamin Shickel} \Email{shickelb@ufl.edu}\\
\Name{Patrick J. Tighe} \Email{ptighe@anest.ufl.edu}\\
\Name{Azra Bihorac} \Email{abihorac@ufl.edu}\\
\Name{Parisa Rashidi} \Email{parisa.rashidi@ufl.edu}\\
\addr University of Florida, Gainesville, FL, United States
}

\begin{document}
\maketitle
\begin{abstract}
Recent deep learning research based on Transformer model architectures has demonstrated state-of-the-art performance across a variety of domains and tasks, mostly within the computer vision and natural language processing domains. While some recent studies have implemented Transformers for clinical tasks using electronic health records data, they are limited in scope, flexibility, and comprehensiveness. In this study, we propose a flexible Transformer-based EHR embedding pipeline and predictive model framework that introduces several novel modifications of existing workflows that capitalize on data attributes unique to the healthcare domain. We showcase the feasibility of our flexible design in a case study in the intensive care unit, where our models accurately predict seven clinical outcomes pertaining to readmission and patient mortality over multiple future time horizons.
\end{abstract}
\begin{keywords}
Transformer, deep learning, electronic health records, mortality prediction
\end{keywords}

\section{Introduction}
\label{sec:introduction}
Through the course of a typical intensive care unit (ICU) admission, a variety of patient-level data is collected and recorded into electronic health records (EHR) systems. Patient data is diverse, including measurements such as vital signs, laboratory tests, medications, and clinician-judged assessment scores. While primarily used for ad-hoc clinical decision-making and administrative tasks such as billing, patient-centric data can also be used to build automated machine learning systems for assessing overall patient health and predicting recovering or worsening patient trajectories.

Patient mortality risk is often used as a proxy for overall ICU patient acuity, both in traditional illness severity scores like SOFA \citep{Vincent1996,Vincent1998} and more recent machine learning approaches such as DeepSOFA \citep{Shickel2019}. Whether manually calculated or algorithmically computed, nearly all of these systems rely on measurements from a set of handpicked clinical descriptors thought to be most indicative of overall patient health. Given the breadth of data available in modern EHR systems, there is untapped potential for enhanced patient modeling contained in the large amount of unused patient data.

Several recent studies have demonstrated the predictive accuracy and patient modeling capacity of deep learning implementations in healthcare, using models such as recurrent neural networks (RNN) \citep{Shickel2019,Choi2015,Choi,Choi2016c,Sha2017,Lipton2015} and convolutional neural networks (CNN) \citep{Lin2019,Nguyen2016}.

Recently, Transformer models \citep{Vaswani2017} have garnered increased attention in the deep learning community due to their state-of-the-art results on a variety of natural language processing (NLP) tasks, particularly when using schemes such as Bidirectional Encoder Representations from Transformers (BERT) \citep{Devlin2018}.

From a temporal perspective, one advantage the Transformer offers is its parallel processing characteristics. Rather than processing data points sequentially, the Transformer views all available data at once, modeling attention-based relationships between all input time steps. In contrast, models such as RNNs require distinct temporal separation within input sequences, and usually demand a regular sample interval between adjacent time steps. As clinical EHR data is recorded at highly irregular frequency and is often missing measurements, a large amount of data preprocessing is typically required in the form of temporal resampling to a fixed frequency, and an imputation scheme to replace missing values. Furthermore, given that several EHR measurements are often recorded at the same timestamp, typical machine learning workflows aggregate temporally adjacent measurements into mean values contained in resampled time step windows, or perform random shuffling procedures before training models. Given its parallel and fundamentally temporally agnostic attributes, the Transformer is capable of distinctly processing all available measurements, even those occurring at the same timestamp. Additionally, the Transformer is able to process whichever data happens to be available, reducing the need for potentially bias-prone techniques to account for data missingness.

In this study, we showcase the feasibility of a highly flexible Transformer-based patient acuity prediction framework in the critical care setting. Our contributions can be summarized by the following:

\begin{itemize} 
\item Our flexible system design incorporates a diverse set of EHR input data that does not require \textit{a priori} identification of clinically relevant input variables, and can work with any data contained in EHR platforms.

\item In contrast to recent Transformer approaches that either use discrete medical concepts \citep{Li2020,Rasmy,Li} or continuous measurements from a handpicked set of features \citep{Song2018}, we introduce a data embedding scheme that jointly captures both concept and corresponding measurement values of a wide variety of disjoint clinical descriptors.

\item In our novel embedding module, we introduce a mechanism for combining both absolute and relative temporality as an improvement over traditional positional encoding. 

\item We present an input data scheme with minimal preprocessing, obfuscating the need for potentially biased temporal resampling or missing value imputation common in many other sequential machine learning approaches.

\item We expand BERT's [CLS] token for classification into several distinct tokens for predicting multiple-horizon patient mortality, ICU readmission, and hospital readmission in a novel multi-task learning environment.

\item Rather than typical concatenation with sequential representation, we incorporate static patient information in a novel way using a global self-attention token so that every sequential time step is compared with the static pre-ICU representation.

\item We show that the Longformer \citep{Peters2019} can be applied to long EHR patient data sequences to minimize required computation while retaining superior performance.

\end{itemize}

\section{Related Work}\label{sec:related_work}
\subsection{Transformer models}
First introduced by Vaswani et al. \citep{Vaswani2017} for machine translation tasks, the Transformer is a deep learning architecture built upon layers of self-attention mechanisms. The Transformer views attention as a function of keys K, queries Q, and values V. In the work of Vaswani et al. \citep{Vaswani2017}, all three elements came from the same input sequence, and is why their style of attention is referred to as self-attention. In a similar manner to previously described works, compatibility between a key and query is used to weight the value, and in the case of self-attention, each element of an input sequence is represented as a contextual sum of the alignment between itself and every other element. Similar to the memory networks of \citet{Sukhbaatar2015}, the Transformer also involves the addition of a positional encoding vector to preserve relative order information between input tokens. 

An end-to-end Transformer architecture typically includes both an encoder and decoder component. While critical for many NLP tasks such as machine translation, our architecture utilizes only the Transformer encoder, which encodes input sequences into hidden representations that are subsequently used for predicting patient mortality.

A comprehensive overview of the Transformer and BERT is beyond the scope of this section; we refer interested readers to Vaswani et al. \citep{Vaswani2017} and Devlin et al. \citep{Devlin2018}, respectively.

Briefly, the first stage of a Transformer encoder typically includes an embedding component, where each input sequence element is converted to a hidden representation that is fed into the remainder of the model. In its original NLP-centered design where inputs are sequences of textual tokens, a traditional embedding lookup table is employed to convert such tokens into continuous representations. Unlike similar sequential models like RNNs or CNNs, the Transformer is fundamentally temporally agnostic and processes all tokens simultaneously rather than sequentially. As such, the Transformer embedding module must inject some notion of temporality into its element embeddings. In typical Transformer implementations, this takes the form of a positional encoding vector, where the position of each element is embedded by sinusoidal lookup tables, which is subsequently added to the token embeddings. The primary aim of such positional embeddings is to allow the model to understand local temporality between nearby sequence elements.

At each layer of a Transformer encoder, a representation of every input sequence element is formed by summing self-attention compatibility scores between the element and every other element in the sequence. Typical with other deep learning architectures, as more layers are added to the encoder, the representations become more abstract.

The recent NLP method BERT \citep{Devlin2018} is based on Transformers, and at present time represent state of the art in a variety of natural language processing tasks. In addition to its novel pretraining scheme, BERT also prepends input sequences with a special [CLS] token before a sequence is passed through the model. The goal of this special token is to capture the combined representation of the entire sequence, and for classification tasks is used for making predictions.

While the Transformer is in one sense more efficient than its sequential counterparts due to its ability to parallelize computations at each layer, one of the main drawbacks is its required memory consumption. Since each input element of a sequence of length $n$ must be compared with every other input element in the sequence, typical Transformer implementations require memory on the order of $O(n^2)$. While acceptable for relatively short sequences, the memory consumption quickly becomes problematic for very long sequences. Decreasing the memory requirement of Transformers is an area of ongoing research.

One potential solution was proposed by \citet{Peters2019} in their Longformer architecture. Rather than computing full $n^2$ self-attentions, they propose a sliding self-attention window of specified width, where each input sequence element is compared only with neighboring sequence elements within the window. They extend this to include user-specified global attention patterns (such as on the special [CLS] tokens for classification) that are always compared with every element in the sequence. Through several NLP experiments, they demonstrate the promising ability of the Longformer to approximate results from a full Transformer model.

\subsection{Transformers in healthcare}

Given the similarity between textual sequences and temporal patient data contained in longitudinal EHR records, several works have begun exploring the efficacy of Transformers and modifications of BERT for clinical applications using electronic health records. In terms of patient data modalities, existing implementations of Transformers in a clinical setting tend to fall under three primary categories:

\paragraph{Clinical text} Perhaps the most aligned with the original BERT implementation, several studies adapt and modify BERT for constructing language models from unstructured text contained in clinical notes. The ClinicalBERT framework of \citet{Huang} used a BERT model for learning continuous representations of clinical notes for predicting 30-day hospital readmission. \citet{Zhang2020} pretrained a BERT model on clinical notes to characterize inherent bias and fairness in clinical language models.

\paragraph{Continuous measurements} \citet{Song2018}'s SAnD architecture developed Transformer models for several clinical prediction tasks using continuous multivariate clinical time series.

\paragraph{Discrete medical codes} The majority of existing EHR Transformer research has focused on temporal sequences of discrete EHR billing codes. \citet{Li}'s BEHRT framework modified the BERT paradigm for predicting future disease from diagnosis codes. Med-BERT \citep{Rasmy} demonstrated the performance advantages of a contextualized clinical pretraining scheme in conjunction with a BERT modification. RAPT \citep{Ren2021} used a modified Transformer pretraining scheme to overcome several challenges with sparse EHR data. SETOR \citep{Peng} utilized neural ordinary differential equations with medical ontologies to construct a Transformer model for predicting future diagnoses. RareBERT \citep{Associates2021} extends Med-BERT for diagnosis of rare diseases. \citet{Meng2021} used Transformers for predicting depression from EHR. Hi-BEHRT \citep{Li} extends BEHRT using a hierarchical design to expand the receptive field to capture longer patient sequences. \citet{Choia} and \citet{Shang2019}'s G-BERT architecture capitalize on the inherent ontological EHR structure.

In contrast to the isolated data modalities implemented in existing EHR Transformers, the novel embedding scheme utilized in our models combines both discrete and continuous patient data to generate a comprehensive representation of distinct clinical events and measurements.

\section{Methods}
\subsection{Cohort}
The University of Florida Integrated Data Repository was used as an honest broker to build a single-center longitudinal dataset from a cohort of adult patients admitted to intensive care units at University of Florida Health between January 1st, 2012 and and September 5th, 2019.

We excluded ICU stays lasting less than 1 hour or more than 10 days. Our final cohort consisted of 73,190 distinct ICU stays from 69,295 hospital admissions and 52,196 unique patients. The median length of stay in the ICU was 2.7 days. 

We divided our total cohort of ICU stays into a development cohort of 60,516 ICU stays (80\%) for training our models, and a validation cohort of 12,674 ICU stays (20\%) for evaluating their predictive performance. 10\% of the development set was used for within-training validation and early stopping. The cohort was split chronologically, where the earliest 80\% of ICU stays was used for training, and the most recent 20\% used for evaluation. Splitting was performed based on the first encounter of each patient to ensure that the same patient was not present in both development and validation sets.

\subsection{Data}\label{sec:data}
We extracted patient data from several EHR data sources: sociodemographics and information available upon hospital admission, summarized patient history, vital signs, laboratory tests, medication administrations, and numerical assessments from a variety of bedside scoring systems. We did not target or manually select any specific ICU variables, instead using all such data contained in our EHR system. A full list of variables used in our experiments is shown in \tableref{tab:list_of_variables}.

\textbf{Static Data}: For each ICU stay, we extracted a set of non-sequential clinical descriptors pertaining to patient characteristics, admission information, and a summarized patient history from the previous year. Patient-level features included several demographic indicators, comorbidities, admission type, and neighborhood characteristics derived from the patient's zip code. Patient history consisted of a variety of medications and laboratory test results up to one year prior to hospital admission (\tableref{tab:list_of_variables}). Historical patient measurement features were derived from a set of statistical summaries for each descriptor (minimum, maximum, mean, standard deviation).

\textbf{Temporal Data}: For each ICU stay, we extracted all available vital signs, laboratory tests, medication administrations, and bedside assessment scores recorded in our EHR system while the patient was in the ICU (\tableref{tab:list_of_variables}). We refer to each extracted measurement as a clinical event. Each event was represented as a vector containing the name of the measurement (e.g. ``noninvasive systolic blood pressure''), the elapsed time from ICU admission, the current measured value, and eight cumulative value-derived features corresponding to prior measurements of the same variable earlier in the ICU stay (mean, median, count, minimum, maximum, standard deviation, first value, elapsed time since most recent measurement). For bedside assessment scores with multiple sub-components, we treated each sub-component as a distinct measurement. Invasive and noninvasive measurements were treated as distinct tokens. We excluded ICU stays with sequence lengths longer than 12,000 tokens, and the resulting mean sequence length in our cohorts was 1,996.
 
\textbf{Data Processing}: Categorical features present in the pre-ICU static data were converted to one-hot vectors and concatenated with the remaining numerical features. Missing static features were imputed with training cohort medians, but no such imputation was required for the tokenized temporal ICU data. Binary indicator masks were computed and concatenated with static features to capture patterns of missingness.

Static features were standardized to zero mean and unit variance based on values from the training set.  For each variable name in the temporal ICU data, corresponding continuous measurement value features were individually standardized in the same manner. ICU measurement timestamps were converted to number of elapsed hours from ICU admission, and were similarly standardized based on training cohort values. 

ICU measurement names were converted to unique integer identifiers in a similar manner to standard tokenization mapping procedures in NLP applications. Each temporal clinical event was also associated with an integer position index. While similar to the positional formulations in NLP applications, we introduce one key distinction that is more suitable for Transformers based on EHR data: we do not enforce the restriction that positional indices are unique, and if two clinical events occurred at the same EHR timestamp, they are associated with the same sequential position index.

Each temporal measurement token consisted of integer positional identifier, integer variable identifier, continuous elapsed time from ICU admission, and eight continuous features extracted from current and prior measurement values.

Following data extraction and processing, each ICU stay was associated with two sets of data: (1) a single vector $x_{static} \in \mathbb{R}^{718 x 1}$ of 718 static pre-ICU features, and (2) a matrix of $T$ temporal ICU measurements $ x_{temporal} \in \mathbb{R}^{T x 12}$ including token position and identifier. Across our entire population, the temporal ICU measurements included 19 unique vital signs, 106 unique laboratory tests, 345 unique medication administrations, and 29 bedside assessment score components; however, each ICU stay only included a subset of such total variables, and its corresponding temporal sequence  only included what was measured during the corresponding ICU stay. One of the benefits of our proposed EHR embedding framework is the lack of resampling, propagation, imputation, or other such temporal preprocessing typically performed in related sequential modeling tasks.

\subsection{Clinical Outcomes}
For each ICU stay, we sought to predict seven clinical outcomes related to patient illness severity: ICU readmission within the same hospital encounter, 30-day hospital readmission, inpatient mortality, 7-day mortality, 30-day mortality, 90-day mortality, and 1-year mortality. Our model is formulated in a multi-task fashion, and simultaneously estimates risk for all seven clinical targets.

\subsection{Model Architecture}
\begin{figure*}[!ht]
\centering
\includegraphics[width=0.9\textwidth]{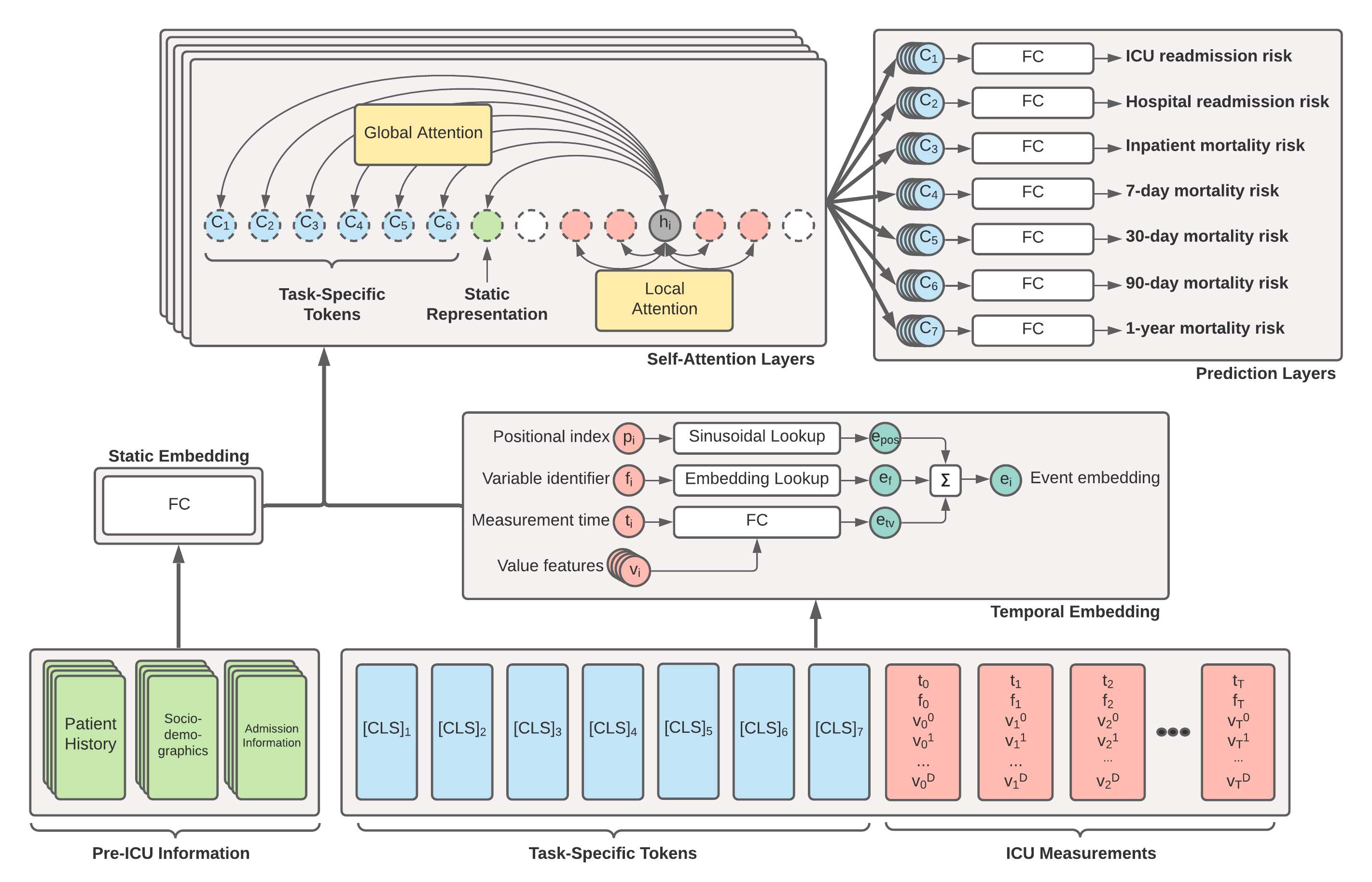}
\label{fig:transformer_model_architecture}
\caption{Overview of our proposed generalized EHR transformer network for simultaneously predicting multiple patient outcomes in the ICU. Pre-ICU information includes summarized history of patient medications and laboratory tests, sociodemographic indicators, and features relating to hospital admission. Temporal ICU measurements take the flexible form of tuples: (p = non-unique positional index of clinical event based on timestamp, t = elapsed time from ICU admission, f = unique measurement identifier integer, $\vec{v}$ = set of continuous features derived from measured values). Task-specific [CLS] tokens are assigned t = time of prediction and $\vec{v}$ = 0. Tokens are individually embedded and passed through a stack of Longformer layers with sliding self-attention windows. Global attention is applied to static feature representation and prediction tokens. The concatenation of each layer's [CLS] representations are used for a given task to predict the desired mortality risk. Not shown: Transformer feedforward network and nonlinear activations. FC: fully-connected layers.}
\end{figure*}

The primary driver behind our model for ICU patient mortality prediction is the Transformer encoder \citep{Vaswani2017}. Our modified model utilizes the global and sliding window mechanism introduced by the Longformer \citep{Peters2019} along with special classification tokens from BERT \citep{Devlin2018}. \figureref{fig:transformer_model_architecture} shows a high-level overview of our Transformer architecture.

 \textbf{Novel Embedding}: In typical Transformer implementations, one-dimensional input sequences consist of integer-identified tokens (such as textual tokens or discrete clinical concepts) that are embedded using a lookup table, after which a positional encoding vector is added to inject local temporality. For existing applications of Transformers with EHR data, the values of a given measurement are not factored into its representation.

Our embedding scheme introduces three novelties that offer improvements for clinical prediction tasks. First, positional indices are derived from EHR record times and are not unique (see \sectionref{sec:data}), allowing for multiple tokens to share the same positional index and resulting positional encoding. Rather than enforce an arbitrary sequence order or implement a random shuffling procedure for simultaneous tokenized events, this modification is more flexible with respect to clinical workflows.

Second, in addition to novel framing of relative and local temporal relationships through positional encoding modifications, each clinical event token also explicitly includes absolute temporality in the form of a feature indicating the elapsed hours from ICU admission. We hypothesized that the injection of both relative and absolute temporality would allow the Transformer to better model patient trajectories.

Finally, each clinical event in our tokenized input sequences consists of several continuous measurement values in addition to the discrete token identifiers (see \sectionref{sec:data}). To our knowledge, no other work integrates both discrete and continuous data in this manner, with the majority of recent research opting for discrete medical codes only (\sectionref{sec:related_work}). We augment discrete variable tokens with continuous measurement values into our embedding to better capture recovery or worsening trends as a patient progresses through an ICU stay.

Our embedding module consists of (1) a traditional lookup table used for measurement name identifier, (2) a sinusoidal positional embedding table, and (3) a single fully-connected layer for embedding absolute time and value-derived features. The final sequence embedding is the summation of three embedded vectors: (1) the embedding of absolute time with corresponding cumulative values, (2) the measurement token identifier embedding, and (3) a traditional sinusoidal positional encoding. In our implementation, the sinusoidal positional encoding is based on the position of unique measurement times in the input sequence: for an example sequence of measurement hours $[0.1, 0.2, 0.2, 0.3, 0.3]$, the positional indices are computed as $[0, 1, 1, 2, 2]$.

\textbf{Novel Multi-Task Global Tokens}:
In the original BERT implementation, a single special [CLS] token is prepended to input sequences that is meant to capture a global representation of the entire sequence. We extend this notion by prepending each sequence with 7 such special tokens: one for each of our clinical outcomes. As each token in our data scheme consists of a (time, name, values) 12-tuple, we set time of each [CLS] token equal to the total ICU length of stay and all values equal to zero. The special token identifiers are embedded in a similar fashion to other ICU measurement tokens. In the Longformer implementation in our encoder, we set each of the multi-task tokens to compute global attention, so that self-attentions are computed among all sequence elements for each clinical outcome token.
 
\textbf{Novel Inclusion of Static Patient Data}: In many sequential models for clinical prediction, a final encounter representation is obtained by concatenating the pre-sequence static patient representation with the sequential representation. In our work, we prepend each ICU sequence with the representation obtained from passing the static patient information vector through a fully-connected network. We assign this static token as global, so that every time step computes attention with the static data. We hypothesized that this more fine-grained injection of patient information at every time step would improve the capacity of our model to learn important and more personalized patient trajectory patterns.

\textbf{Model Details}: Our final model consisted of an embedding layer, followed by 8 Longformer layers, and a separate linear prediction layer for each of our 7 clinical outcomes. For making a task-specific prediction, the task-specific linear layer uses the concatenation of representations corresponding to its special [CLS] token at each of the 8 layers. In our initial Longformer implementation, we used a hidden size of 128, a feedforward size of 512, 8 attention heads, a sliding window of size 128, dropout of 0.1, and a batch size of 21. Hyperparameters were chosen with respect to hardware constraints; hyperparameter optimization will be a focus of future work.

\textbf{Experiment Details}: Models were trained using a development set of 60,516 ICU stays corresponding to 80\% of our total ICU cohort. 10\% of this development set was used for early stopping based on the mean AUROC among all seven clinical outcomes and a patience of four epochs. In this feasibility study, we compared performance against four other ICU prediction models:
\begin{itemize}
	\item Transformer using tokenized data sequences with only discrete code identifiers. In this variant of our proposed framework, we do not include the continuous measurement values in the representation of each event token.
	\item Recurrent neural network (RNN) with gated recurrent units (GRU) using continuous multivariate time series inputs. In this experiment, the flexibility of our tokenization scheme is removed, and more traditional ``tabularized'' input data sequences were constructed where each variable is assigned a distinct column. Sequences were constructed with continuous current values and resampled to 1-hour frequency to align with common practice found in literature. Multi-task predictions were drawn from the final hidden state of the GRU encoder. Static patient information was concatenated with the sequence representation and fed through fully-connected layers before classification.
	\item GRU with attention mechanism. This variant is identical to the above, but with the addition of a simple attention mechanism over the hidden states of the GRU. States are weighted by alignment scores and summed to yield a final attention-based sequential representation.
	\item Tokenized GRU with attention. In this final experimental setting, we used the same novel EHR embedding and tokenization approach as with our Transformer model architecture (see \sectionref{sec:data}), but instead use a GRU with attention mechanism in place of the Transformer model.
\end{itemize}

\section{Results}

\begin{table*}
\label{tab:results}
\centering
\small
\caption{Multi-task prediction results expressed as area under the receiver operating characteristic curve (AUROC).}
\begin{tabular}{p{1.5cm}p{3cm}llllllll}
\small
\multirow{2}{*}{\textbf{Model}} & \multirow{2}{*}{\textbf{Data}} & \multirow{2}{*}{\textbf{Mean}} & \multicolumn{2}{c}{\textbf{Readmission}} & \multicolumn{5}{c}{\textbf{Mortality}}\\
\cmidrule(lr){4-5}
\cmidrule(lr){6-10}
& & & ICU & Hospital & Inpatient & 7-Day & 30-Day & 90-Day & 1-Year\\
\toprule

Transformer & Tokenized Events (Discrete only) & 0.734 & 0.512 & 0.501 & 0.889 & 0.900 & 0.831 & 0.777 & 0.727\\
\midrule
Transformer & Tokenized Events + Continuous Measurement Values & \textbf{0.895} & \textbf{0.843}	 & \textbf{0.697} & \textbf{0.978}	 & \textbf{0.983} & 0.953 & 0.923 & \textbf{0.892}\\
\midrule
GRU & Resampled Multivariate Time Series & 0.869	 & 0.75 & 0.686 & 0.960 & 0.972 & 0.938 & 0.907 & 0.872\\
\midrule
GRU with Attention & Resampled Multivariate Time Series & 0.877 & 0.770 & 0.684 & 0.965 & 0.975 & 0.946 & 0.914 & 0.882\\
\midrule
GRU with Attention & Tokenized Events + Continuous Measurement Values & 0.892 & 0.831 & 0.688 & 0.977 & 0.982 & \textbf{0.954} & \textbf{0.925} & 0.891\\
\bottomrule
\end{tabular}
\end{table*}

At present time, the primary aim of our novel mortality prediction model is not to show state-of-the-art improvements in model accuracy; rather, we present this work as a feasibility study for future research. We believe our novel modifications of existing Transformer architectures for use in clinical EHR applications will result in highly flexible and more personalized patient representations and predictions across a variety of clinical tasks.

In this iteration of our experiments, we did not perform any hyperparameter optimization, instead choosing sensible settings that both highlight the novel aspects of the architecture and work with our hardware constraints. In passing, we note that often parameter tuning is an essential component of enhancing performance, and future iterations of this work will focus on optimizing crucial parameters such as learning rate, dropout, number of self-attention heads, number of self-attention layers, hidden dimension, and size of the sliding self-attention window.

Our preliminary results are shown in \tableref{tab:results}. Our Transformer architecture with novel EHR embedding and tokenization scheme yielded superior mean AUROC (0.895) across all seven clinical prediction tasks, with individual task AUROC ranging from 0.697 (30-day hospital readmission) to 0.983 (7-day mortality). The Transformer using tokenized embeddings that omit continuous measurement values resulted in the lowest mean AUROC (0.734) and worst performance across each of the clinical outcomes, ranging from 0.501 (30-day hospital readmission) to 0.900 (7-day mortality). 

In terms of GRU baseline models, the traditional model and data processing scheme resulted in the worst baseline performance, with mean AUROC of 0.869 and task AUROC ranging from 0.686 (30-day hospital readmission) to 0.972 (7-day mortality). The augmentation of this model and data scheme with an simple attention mechanism improved the performance to a mean AUROC of 0.877.

The best GRU baseline model used our novel EHR embedding, tokenization, and representation pipeline. This model yielded a mean AUROC of 0.892 with individual task AUROC ranging from 0.688 to 0.982. It performed best for predicting 30-day. mortality and 90-day mortality, although the difference between the Transformer is minimal.

Across all models and data representation schema, 30-day hospital readmission proved the most difficult task, followed by ICU readmission. Among the multiple prediction horizons for patient mortality, models were best able to predict 7-day mortality, followed by inpatient mortality, 30-day mortality, 90-day mortality, and 1-year mortality.

\section{Discussion}
This work presents a novel ICU acuity estimation model inspired by recent breakthroughs in Transformer architectures. Our proposed model framework incorporates several novel modifications to the existing Transformer architecture that make it more suitable for processing EHR data of varying modalities. Through initial feasibility experiments, our model outperformed common variants of RNN baselines, and we feel our approach holds promise for incorporating additional EHR-related outcome prediction tasks and additional sources of EHR input data. Future work will include tuning hyperparameters with a focus on predictive performance, and examining the self-attention compatibility scores between elements and task tokens for improving clinician interpretability and identifying important contributing descriptors for each task.

One of the advantages of our work is that input elements are treated as distinct. For example, if heart rate, respiratory rate, and SPO2 were recorded at the same timestamp in an EHR system, our framework operates on these individual elements, rather than combining them into a single aggregated time step as in similar RNN or CNN-based work. From an interpretability standpoint, combined with the inherent self-attention mechanisms of the Transformer, isolation of inputs allows for improved clarity with respect to important or contributing clinical factors. While one area of recent sequential interpretability research involves multivariate attribution for aggregated time steps \citep{Choi,Qin2017}, Transformer-based approaches such as ours obfuscate the need for multivariate attribution, as attentional alignment scores are assigned to individual measurements. This property highlights the potential for EHR Transformers to shed increased transparency and understanding for clinical prediction tasks built upon complex human physiology.

Furthermore, while many sequential applications of deep learning to EHR (including recent implementations of Transformer techniques) make use only of discrete clinical concepts, our proposed framework extends the representational capacity by integrating continuous measurement values alongside these discrete codes and events. The inclusion of continuous measurement values represents an important step forward, as the measured result of a clinical test or assessment can provide crucial information alongside a simple presence indicator that can help complex models develop a better understanding of patient state and overall health trajectory.

Given the flexible nature of our Transformer framework, each patient input sequence only contains the measurements that were made during the ICU encounter. The advantages for EHR applications are twofold. First, in traditional RNN or CNN-based work, the distance between time steps is assumed to be fixed, and this is typically achieved by resampling input sequences to a fixed frequency by aggregating measurements within resampled windows, and propagating or imputing values into windows without present values. Such a scheme has the potential for introducing bias, and when using our novel EHR embedding paradigm and Transformer-based modeling approach, the problem of missing values is made redundant given the explicit integration of both absolute and relative temporality for each irregularly measured clinical event. Additionally, in typical deep sequential applications using EHR data, the number of input features at each time step must be constant. This is achieved by an \textit{a priori} identification and extraction of a subset of clinical descriptors thought to be relevant indicators for a given prediction task. As we have shown, when using a Transformer-based approach with our flexible tokenization scheme, any and all EHR measurements can be easily incorporated into the prediction framework, even when some types do not exist for a given patient or ICU encounter, and do not necessitate bias-prone imputation techniques.

In our approach, we introduced a novel method for incorporating static, pre-sequential patient information and patient history into the overall prediction model. Typically, such static information is concatenated with a final sequential representation before making a prediction. We instead include static information as a distinct token in the input sequence, and assign global attention using the Longformer self-attention patterns. In effect, static patient-level information is injected into the self-attention representation of every ICU measurement, allowing more fine-grained and personalized incorporation of changes in overall patient health trajectories.

Another novel contribution we feel can be applied to even non-EHR tasks is the expansion of the special BERT classification token into a separate token per classification target in a multi-task prediction setting. Given the global self-attention patterns between all task tokens and every sequential input element, such a scheme allows the model to develop task-specific data representations that can additionally learn from each other.

This feasibility study has several limitations and is intended as a methodological guiding framework for future multimodal and multi-task EHR Transformer research. Our retrospective dataset is limited to patients from a single-center cohort. We also present results with parameters that maximize our limited hardware capacity; future work will focus on several hyperparameter tuning and model selection procedures. The baseline models we present for comparison are drawn from simplified implementations found in clinical deep learning research, and more recent approaches may offer enhanced predictive performance. From the results in \tableref{tab:results}, one might conclude that our EHR embedding procedure had a larger impact than use of the Transformer architecture, given the competitive AUROC of the attentional GRU baseline when implementing our tokenization pipeline for estimating risk of patient mortality. Future work will focus on disentangling the relative impacts of both model and data representation designs. Given the overall superior performance of the Transformer in conjunction with our flexible EHR embedding techniques, we believe there is great potential for multi-modal patient monitoring using flexible EHR frameworks such as ours. Future research will also focus on augmenting our multi-modal datasets with additional clinical data modalities such as clinical text and images, and pre-training our Transformer architectures with self-supervised prediction schemes across a variety of input data and clinical outcomes.

We feel there is still great potential for exploring additional benefits of our approach with diverse EHR data for a variety of clinical modeling and prediction tasks, especially in the realm of clinical interpretability. Since Transformers are fundamentally composed of attention mechanisms, they can be analyzed with respect to particular outcomes, time points, or variables of interest to highlight important contributing factors to overall risk estimation. Future research will place heavy emphasis on analyses of self-attention distributions between input variables and clinical outcomes to further the clinical explainability and enhance the clinical trust of Transformers in healthcare.

\bibliography{references.bib}

\section{Acknowledgements}
PR was supported by National Science Foundation CAREER award 1750192, 1R01EB029699 and 1R21EB027344 from the National Institute of Biomedical Imaging and Bioengineering (NIH/NIBIB), R01GM-110240 from the National Institute of General Medical Science (NIH/NIGMS), 1R01NS120924 from the National Institute of Neurological Disorders and Stroke (NIH/NINDS), and by R01 DK121730 from the National Institute of Diabetes and Digestive and Kidney Diseases (NIH/NIDDK).

A.B. was supported R01 GM110240 from the National Institute of General Medical Sciences (NIH/NIGMS), R01 EB029699 and R21 EB027344 from the National Institute of Biomedical Imaging and Bioengineering (NIH/NIBIB), R01 NS120924 from the National Institute of Neurological Disorders and Stroke (NIH/NINDS), and by R01 DK121730 from the National Institute of Diabetes and Digestive and Kidney Diseases (NIH/NIDDK).

The content is solely the responsibility of the authors and does not necessarily represent the official views of the National Institutes of Health.

\appendix

\section{Supplemental Tables}\label{apd:first}
\begin{table*}
\centering
\small
\caption{Summary statistics for experimental ICU cohorts.}
\begin{tabular}{lll}
\small
	& Development Cohort  & Validation Cohort \\
& (n = 60,516) & (n = 12,674)\\
\toprule
Patients, n & 41,881 & 10,315\\
Hospital encounters, n & 57,168 & 12,127\\
Age, years, median (25th, 75th) & 61.0 (49.0, 71.0) & 62.0 (49.0, 73.0)\\
Female, n (\%) & 27,380 (45.2) & 5,616 (44.3)\\
Body mass index, median (25th, 75th) & 26.9 (23.0, 32.0) & 27.3 (23.3, 32.2)\\
Hospital length of stay, days, median (25th, 75th) & 6.7 (3.6, 12.1) & 6.4 (3.3, 11.5)\\
ICU length of stay, days, median (25th, 75th) & 2.8 (1.5, 5.1) & 2.9 (1.6, 5.5)\\
Time to hospital discharge, days, median (25th, 75th) & 1.9 (0.0, 4.8) & 1.1 (0.0, 4.1)\\
Hispanic, n (\%) & 2,130 (3.5) & 539 (4.3)\\
Non-English speaking, n (\%) & 1,092 (1.8) & 233 (1.8)\\

Marital status, n (\%) & &\\
\quad Married & 26,084 (43.1) & 5,457 (43.1)\\
\quad Single & 21,844 (36.1) & 4,931 (38.9)\\
\quad Divorced & 11,905 (19.7) & 2,142 (16.9)\\

Smoking status, n (\%) & &\\
\quad Never & 20,180 (33.3) & 4,653 (36.7)\\
\quad Former & 19,378 (32.0) & 4,167 (32.9)\\
\quad Current & 12,094 (20.0) & 2,326 (18.4)\\

Insurance status, n (\%) & &\\
\quad Medicare & 31,447 (52.0) & 6,543 (51.6)\\
\quad Private & 13,115 (21.7) & 2,912 (23.0)\\
\quad Medicaid & 10,208 (16.9) & 1,999 (15.8)\\
\quad Uninsured & 5,746 (9.5) & 1,220 (9.6)\\

Comorbidities, n (\%) & &\\
\quad Charlson comorbidity index, median (25th, 75th) & 2.0 (0.0, 4.0) & 2.0 (0.0, 4.0)\\
\quad Myocardial infarction & 7,537 (12.5)	& 1,985 (15.7)\\
\quad Congestive heart failure	 & 14,897 (24.6)	& 3,380 (26.7)\\
\quad Peripheral vascular disease & 10,005 (16.5) & 2,185 (17.2)\\
\quad Cerebrovascular disease	& 8,981 (14.8) & 1,720 (13.6)\\
\quad Chronic pulmonary disease & 17,938 (29.6) & 	3,473 (27.4)\\
\quad Metastatic carcinoma & 3,377 (5.6) & 812 (6.4)\\
\quad Cancer & 8202 (13.6) & 1,808 (14.3)\\
\quad Mild liver disease & 4,745 (7.8) & 960 (7.6)\\
\quad Moderate/severe liver disease & 1,856 (3.1) & 374 (3.0)\\
\quad Diabetes without complications & 14,137 (23.4) & 2,395 (18.9)\\
\quad Diabetes with complications & 5,052 (8.3) & 1,736 (13.7)\\
\quad AIDS & 442 (0.7) & 53 (0.4)\\
\quad Dementia & 1,692 (2.8) & 559 (4.4)\\
\quad Paraplegia/hemiplegia & 3,465 (5.7) & 769 (6.1)\\
\quad Peptic ulcer disease & 1,110 (1.8) & 187 (1.5)\\
\quad Renal disease & 11,878 (19.6)	 & 2,493 (19.7)\\
\quad Rheumatologic disease & 1,794 (3.0) & 342 (2.7)\\

Neighborhood characteristics, median (25th, 75th) & &\\
\quad Total population, n $\times 10^3$  & 17.0 (10.6, 26.4) & 17.6 (10.6, 26.7)\\
\quad Distance to hospital, km & 39.3 (17.9, 69.1)	 & 42.4 (20.2, 76.5)\\
\quad Median income, dollars $\times 10^3$ & 40.1 (33.8, 46.7) & 40.1 (35.1, 47.4)\\
\quad Poverty rate, \% & 19.6 (14.0, 27.7)	 & 19.3 (13.7, 26.7)\\
\quad Rural area, n & 22543 (37.3) & 	4691 (37.0)\\

Clinical outcomes, n (\%) & &\\
\quad ICU readmission before hospital discharge & 3,583 (5.9) & 613 (4.8)\\
\quad 30-day hospital readmission & 14,221 (23.5) & 1,325 (10.5)\\
\quad Inpatient mortality & 5,813 (9.6) & 1,131 (8.9)\\
\quad 7-day mortality & 5,237 (8.7) & 1,022 (8.1)\\
\quad 30-day mortality & 7,056 (11.7) & 1,380 (10.9)\\
\quad 90-day mortality & 9,197 (15.2) & 1,785 (14.1)\\
\quad 1-year mortality & 12,991 (21.5) & 2,288 (18.1)\\
\bottomrule
\end{tabular}
\end{table*}

\clearpage
\onecolumn
\begin{longtable}{ll}
\caption{Summary of variables used in Transformer experiments.}\\

    \toprule
    \textbf{Variable} & \textbf{Type}\\
    \midrule
    \endfirsthead
    \caption[]{Continued.}\\
    \toprule
    \textbf{Variable} & \textbf{Type}\\
    \midrule
    \endhead
    \endfoot
    \bottomrule
    \endlastfoot
    
    \textbf{Patient Demographics} & \\
    \quad Age & Static\\
    \quad Sex & Static\\
    \quad Ethnicity & Static\\
    \quad Race & Static\\
    \quad Language & Static\\
    \quad Marital Status & Static\\
    \quad Smoking Status & Static\\
    \quad Insurance Provider & Static\\
    
    \textbf{Patient Residential Information} & \\
    \quad Total Population & Static\\
    \quad Distance from Hospital & Static\\
    \quad Rural/Urban & Static\\
    \quad Median Income & Static\\
    \quad Proportion Black & Static\\
    \quad Proportion Hispanic & Static\\
    \quad Percent Below Poverty Line & Static\\
    
    \textbf{Patient Admission Information} & \\
    \quad Height & Static\\
    \quad Weight & Static\\
    \quad Body Mass Index & Static\\
    \quad 17 Comorbidities Present at Admission & Static\\
    \quad Charlson Comorbidity Index & Static\\
    \quad Presence of Chronic Kidney Disease & Static\\
    \quad Admission Type& Static\\
    
    \multicolumn{2}{l}{\textbf{Patient History: Medications $^a$}}\\
    \quad ACE Inhibitors & Static\\
    \quad Aminoglycosides & Static\\
    \quad Antiemetics & Static\\
    \quad Aspirin & Static\\
    \quad Beta Blockers & Static\\
    \quad Bicarbonates & Static\\
    \quad Corticosteroids & Static\\
    \quad Diuretics & Static\\
    \quad NSAIDS & Static\\
    \quad Vasopressors/Inotropes & Static\\
    \quad Statins & Static\\
    \quad Vancomycin & Static\\
    \quad Nephrotoxic Drugs & Static\\
    
    \multicolumn{2}{l}{\textbf{Patient History: Laboratory Test Results $^b$}}\\
    \quad Serum Hemoglobin & Static\\
    \quad Urine Hemoglobin & Static\\
    \quad Serum Glucose & Static\\
    \quad Urine Glucose & Static\\
    \quad Urine Red Blood Cells & Static\\
    \quad Urine Protein & Static\\
    \quad Serum Urea Nitrogen & Static\\
    \quad Serum Creatinine & Static\\
    \quad Serum Calcium & Static\\
    \quad Serum Sodium & Static\\
    \quad Serum Potassium & Static\\
    \quad Serum Chloride & Static\\
    \quad Serum Carbon Dioxide & Static\\
    \quad White Blood Cells & Static\\
    \quad Mean Corpuscular Volume & Static\\
    \quad Mean Corpuscular Hemoglobin & Static\\
    \quad Hemoglobin Concentration & Static\\
    \quad Red Blood Cell Distribution & Static\\
    \quad Platelets & Static\\
    \quad Mean Platelet Volume & Static\\
    \quad Serum Anion Gap & Static\\
    \quad Blood pH & Static\\
    \quad Serum Oxygen & Static\\
    \quad Bicarbonate & Static\\
    \quad Base Deficit & Static\\
    \quad Oxygen Saturation & Static\\
    \quad Band Count & Static\\
    \quad Bilirubin & Static\\
    \quad C-Reactive Protein & Static\\
    \quad Erythrocyte Sedimentation Rate & Static\\
    \quad Lactate & Static\\
    \quad Troponin T/I & Static\\
    \quad Albumin & Static\\
    \quad Alaninen & Static\\
    \quad Asparaten & Static\\
    
    \multicolumn{2}{l}{\textbf{ICU Vital Signs}}\\
	\quad Systolic Blood Pressure $^c$ & Temporal\\
	\quad Diastolic Blood Pressure $^c$ & Temporal\\
	\quad Mean Arterial Pressure $^c$ & Temporal\\
	\quad Heart Rate & Temporal\\
	\quad Respiratory Rate & Temporal\\
	\quad Oxygen Flow Rate & Temporal\\
	\quad Fraction of Inspired Oxygen (FIO2) & Temporal\\
	\quad Oxygen Saturation (SPO2) & Temporal\\
	\quad End-Tidal Carbon Dioxide (ETCO2) & Temporal\\
	\quad Minimum Alveolar Concentration (MAC) & Temporal\\
	\quad Positive End-Expiratory Pressure (PEEP) & Temporal\\
	\quad Peak Inspiratory Pressure (PIP) & Temporal\\
	\quad Tidal Volume & Temporal\\
	\quad Temperature & Temporal\\
	
	\multicolumn{2}{l}{\textbf{ICU Assessment Scores $^d$}}\\
	\quad ASA Physical Status Classification & Temporal\\
	\quad Braden Scale & Temporal\\
	\quad Confusion Assessment Method (CAM) & Temporal\\
	\quad Modified Early Warning Score (MEWS) & Temporal\\
	\quad Morse Fall Scale (MFS) & Temporal\\
	\quad Pain Score & Temporal\\
	\quad Richmond Agitation-Sedation Scale (RASS) & Temporal\\
	\quad Sequential Organ Failure Assessment (SOFA) & Temporal\\
	
	\multicolumn{2}{l}{\textbf{ICU Laboratory Tests $^e$}}\\
	\quad 106 distinct lab tests present in EHR system & Temporal\\
	
	\multicolumn{2}{l}{\textbf{ICU Medications $^e$}}\\
	\quad 345 distinct medications present in EHR system & Temporal\\
	
\label{tab:list_of_variables}
  
\end{longtable}
$^a$ Extracted features included total counts of administered medications up to one year prior to hospital admission.

$^b$ Extracted features included total counts of recorded laboratory test results and minimum, maximum, mean, and standard deviation of measurement values up to one year prior to hospital admission. Both serum and urine-based tests extracted separately when available.

$^c$ Invasive and non-invasive readings for systolic blood pressure, diastolic blood pressure, and mean arterial pressure were treated as distinct event tokens.

$^d$ For assessment scores with multiple sub-components, each component was treated as a distinct timestamped measurement, resulting in 30 such assessment measurements.

$^e$ We retained distinct laboratory tests and medications that were administered in at least 1\% of the training cohort of ICU stays.
\end{document}